# BIAS: Transparent reporting of biomedical image analysis challenges

Lena Maier-Hein
*Division of Computer Assisted Medical Interventions (CAMI), German Cancer Research Center (DKFZ)*
l.maier-hein@dkfz-heidelberg.de

Anne L. Martel
*Physical Sciences, Sunnybrook Research Institute*

*Department Medical Biophysics, University of Toronto*

Pierre Jannin
*Laboratoire Traitement du Signal et de l'Image (LTSI) - UMR_S 1099, Université de Rennes 1, Inserm*

Sinan Onogur
*Division of Computer Assisted Medical Interventions (CAMI), German Cancer Research Center (DKFZ)*

Annette Kopp-Schneider
*Division of Biostatistics, German Cancer Research Center (DKFZ)*

Annika Reinke
*Division of Computer Assisted Medical Interventions (CAMI), German Cancer Research Center (DKFZ)*

Tal Arbel
*Centre for Intelligent Machines, McGill University*

Allan Hanbury
*Institute of Information Systems Engineering, Technische Universität (TU) Wien*

*Complexity Science Hub Vienna, Vienna*

Julio Saez-Rodriguez
*Institute of Computational Biomedicine, Heidelberg University*

*Faculty of Medicine, Heidelberg University Hospital*

*Joint Research Centre for Computational Biomedicine, Rheinisch-Westfälische Technische Hochschule (RWTH) Aachen*

Michal Kozubek
*Centre for Biomedical Image Analysis, Masaryk University*

Matthias Eisenmann
*Division of Computer Assisted Medical Interventions (CAMI), German Cancer Research Center (DKFZ)*

Henning Müller
*University of Applied Sciences Western Switzerland (HES-SO)*

*Medical Faculty, University of Geneva*

Bram van Ginneken
*Department of Radiology and Nuclear Medicine, Medical Image Analysis, Radboud University Centre*

Bennett Landman
*Electrical Engineering, Vanderbilt University, Nashville, Tennessee*

*Abstract*— The number of biomedical image analysis challenges organized per year is steadily increasing. These international competitions have the purpose of benchmarking algorithms on common data sets, typically to identify the best method for a given problem. Recent research, however, revealed that common practice related to challenge reporting does not allow for adequate interpretation and reproducibility of results. To address the discrepancy between the impact of challenges and the quality (control), the Biomedical I mage Analysis ChallengeS (BIAS) initiative developed a set of recommendations for the reporting of challenges. The BIAS statement aims to improve the transparency of the reporting of a biomedical image analysis challenge regardless of field of application, image modality or task category assessed. This article describes how the BIAS statement was developed and presents a checklist which authors of biomedical image analysis challenges are encouraged to include in their submission when giving a paper on a challenge into review. The purpose of the checklist is to standardize and facilitate the review process and raise interpretability and reproducibility of challenge results by making relevant information explicit.

## I. Introduction

The importance of data science techniques in almost all fields of biomedicine is increasing at an enormous pace [1,2]. This holds particularly true for the field of biomedical image analysis, which plays a crucial role in many areas including tumor detection, classification, staging and progression modeling [3,4,5] as well as automated analysis of cancer cell images acquired using microscopy [6,7,8].

While clinical trials are the state of the art methods to assess the effect of new medication in a comparative manner [9], benchmarking in the field of image analysis is performed by so-called *challenges*. Challenges are international competitions, typically hosted by individual researchers, institutes, or societies, that aim to assess the performance of multiple algorithms on identical data sets and encourage benchmarking [10]. They are often published in prestigious journals [11,12,13,14,15,16], are associated with significant amounts of prize money (up to €1 million on platforms like Kaggle [17]) and receive a huge amount of attention, indicated by the number of downloads, citations and views. A recent comprehensive analysis of biomedical image analysis challenges, however, revealed a huge discrepancy between the impact of a challenge and the quality (control) of the design and reporting standard. It was shown that (1) "common practice related to challenge reporting is poor and does not allow for adequate interpretation and reproducibility of results", (2) "challenge design is very heterogeneous and lacks common standards, although these are requested by the community" and (3) "challenge rankings are sensitive to a range of challenge design parameters, such as the metric variant applied, the type of test case aggregation performed and the observer annotating the data" [18]. The authors conclude that "journal editors and reviewers should provide motivation to raise challenge quality by establishing a rigorous review process."

The Enhancing the QUAlity and Transparency Of health Research (EQUATOR) network is a global initiative with the aim of improving the quality of research publications and research itself. A key mission in this context is to achieve accurate, complete and transparent reporting of health research studies to support reproducibility and usefulness. A core activity of the network is to assist in the development, dissemination and implementation of robust reporting guidelines, where a guideline is defined as "a checklist, flow diagram or structured text to guide authors in reporting a specific type of research" [19]. Between 2006 and 2019, more than 400 reporting guidelines have been published under the umbrella of the equator network. A well-known guideline is the CONSORT statement [20,21] developed for reporting of randomized controlled trials. Prominent journals, such as Lancet, JAMA or the British Medical Journal require the CONSORT checklist to be submitted along with the actual paper when reporting results of a randomized controlled trial.

Inspired by this success story, the **B**iomedical **I**mage **A**nalysis Challenge**S** (BIAS) initiative was founded by the challenge working group of the Medical Image Computing and Computer Assisted Intervention (MICCAI) Society board with the goal of bringing biomedical image analysis challenges to the next level of quality.

As a first step towards better scientific practice, this paper of the initiative presents a guideline to standardize and facilitate the writing and reviewing process of biomedical image analysis challenges and help readers of challenges interpret and reproduce results by making relevant information explicit.

Please note that we do not want to put unnecessary restrictions on researchers. For this reason, the template for challenge papers as proposed in the following sections merely serves as guidance, and authors are free to arrange the relevant information in any way they want. What we regard as important is for the information in the paper to be complete, such that transparency and reproducibility can be guaranteed. For this reason, we encourage authors of challenge papers to submit the checklist presented in this manuscript (*Appendix A*, [22]) along with their paper such that reviewers can easily verify whether the information on challenge design and results is comprehensive. If information is missing (represented by "n/a" in the column *reported on page No* of the checklist) it is up to the reviewers to request adding it.

Section 2 introduces the terminology used to describe challenges and describes the process applied to generate this guideline document. Section 3 gives recommendations on how to report the design and results of a biomedical image analysis challenge. The paper then closes with a brief discussion in section 4.

## II. METHODS

In this paper, we define a *biomedical image analysis challenge* as an open competition on a specific scientific problem in the field of biomedical image analysis [18]. A challenge may encompass multiple competitions related to multiple *tasks*, whose participating teams may differ and for which separate rankings/leaderboards/results are generated. For example, a challenge may target the problem of anatomical structure segmentation in computed tomography (CT) images, where one task may refer to the segmentation of the liver and a second task may refer to the segmentation of the kidney. We use the term *case* to refer to a data set for which a participating algorithm produce one result (e.g. a segmentation or classification). Each case must include at least one image of a biomedical imaging modality.

*Metrics* are used to compute the performance of an algorithm for a given case and should reflect the property(ies) of the algorithms to be optimized. Note that we do not use the term *metric* in the strict mathematical sense. Metrics are usually computed by comparing the results of the participating team with a *reference* annotation. We prefer the term *reference* (alternatively: gold standard) compared to *ground truth* because reference annotations are typically only approximations of the (forever unknown) truth [23].

Typically, a challenge has a *training phase* of several weeks or months, at the beginning of which the challenge organizers release training cases with corresponding reference annotations. These annotations help the participating teams develop their method (e.g. by training a machine learning algorithm). Alternatively, the training data are not directly released but participating teams may submit their algorithms to the challenge platform (using Docker containers, for example [24]). Note that the official training phase may be preceded by a *dry run phase*. During this phase, the challenge organizers may themselves work with the data to determine the level of difficulty of the task(s), for example. In the *test phase*, participating teams either upload their algorithms, or they get access to the test cases without the reference annotations and submit the results of their algorithms on the test cases to the challenge organizers. This procedure may be replaced or complemented by an on-site challenge event in which participating teams receive a set of test cases and are asked to produce the corresponding results on-site (typically on the same day).

For many challenges a ranking of the participating teams is produced based on the metric values computed for the test cases. Note that some challenges additionally include a validation phase between training and test phase, in which initial rankings (so-called *leaderboards*) are generated to show participating teams how well their methods generalize. Insights in this step may be used for final parameter tuning. A glossary of some of the terms used in this paper is provided in *Appendix B*.

The procedure to generate this guideline document was heavily based on a previous study related to the critical analysis of common practice in challenge organization [18] and is summarized in the following paragraphs:

**Challenge capture** To analyze the state of the art in the field, the publicly available data on biomedical image analysis challenges was acquired. To capture a challenge in a structured manner, a list of 53 challenge parameters was compiled by a group of 49 scientists from 30 institutions worldwide. These parameters include information on the challenge organization and participation conditions, the mission of the challenge, the challenge data sets (e.g. number of training/test cases, information on imaging protocols), the assessment method (e.g. metrics and ranking scheme) and challenge outcome (e.g. rankings). Analysis of websites hosting and presenting challenges, such as grand-challenge.org, dreamchallenges.org and kaggle.com yielded a list of 150 biomedical image analysis challenges with 549 tasks performed in a time span of 12 years [18]. Between 2004 and 2016, most challenges were organized in the scope of international conferences, primarily the MICCAI conference (48%) and the International Symposium on Biomedical Imaging (ISBI) (24%). More recently, an increasing number of challenges are hosted on platforms like Kaggle [17], Synapse (for the DREAM challenges [25,26]) and crowdAI [27] (for the ImageCLEF challenges). Details on the challenge characteristics (e.g. imaging modalities applied, algorithm categories investigated, number of training/test cases) can be found in [18].

**Analysis of challenge reporting** It was found that reports on biomedical challenges covered only a median of 62% of the 53 challenge parameters identified as relevant by the international consortium. The list of parameters often not reported include some that are crucial for interpretation of results such as information on how metrics are aggregated to obtain a ranking, whether training data provided by challenge organizers may have been supplemented by other data, and how the reference annotation was performed and by whom. It was further found that challenge design is highly heterogeneous, as detailed in [18].

**Prospective structured challenge capture** To address some of the issues, a key conclusion of [18] was to publish the complete challenge design before the challenge by instantiating the list of parameters proposed. To test the applicability of this recommendation, the MICCAI board challenge working group initiated the usage of the parameter list for structured submission of challenge proposals for the MICCAI conferences 2018 and 2019. The submission system required a potential MICCAI 2018/2019 challenge organizer to instantiate at least 90% of a reduced set of 40 parameters (cf. Tab. 1 in [1]) that were regarded as essential for judging the quality of a challenge design proposal. The median percentage of parameters instantiated was 100% (min: 94%) (16/25 submitted challenges in 2018/2019).

**Finalization of checklist** Based on the prospective challenge capture, the parameter list was revised by the MICCAI board challenge working group to improve clarity. A questionnaire was then sent to all co-authors to acquire final feedback on the parameters. Each author had to independently assess every single parameter ($n = 48$) of the list by answering the following questions:
1. I agree with the name (yes/sort of/no).
2. I agree with the explanation (yes/sort of/no).
3. If you do not agree with the name or the explanation, please provide constructive feedback.
4. Please rate the importance of the checklist item. If you think that it is absolutely essential for challenge result interpretation and/or challenge participation put *absolutely essential*. Otherwise choose between *should be included* and *may be omitted*.
5. Please indicate whether the checklist item(s) is (are) essential for challenge review (yes/no).

To identify missing information, participants were also asked to add further relevant checklist items that were not covered and to add any other issue important to compile the checklist. The MICCAI board challenge working group then developed a proposal to address all the comments and points of criticism raised in the poll. In a final conference call with the co-authors of this paper, remaining conflicts were resolved, and the checklist was finalized, resulting in a list of 42 main parameters and 79 sub-parameters.

The following sections describe the authors' recommendations on how to report the design and outcome of individual tasks of a biomedical image analysis challenge based on this parameter list. The corresponding reviewer checklist is provided in *Appendix A* and was uploaded to Zenodo to ensure version control [22].

III. GUIDELINE FOR CHALLENGE REPORTING

Following standard scientific writing guidelines, we propose dividing a challenge paper into the sections *Introduction, Methods, Results* and *Discussion*, where the *Methods* section corresponds to the challenge design and the *Results* section corresponds to the challenge outcome. These sections are preceded by a concise title and abstract as well as a list of representative keywords to summarize the challenge mission and outcome. The following sections give basic recommendations on how to structure and write the individual sections. Appendix A serves as a structured summary of this section.

*A. Title, Abstract and Keywords*

The *title* should convey the essential information on the challenge mission. In particular, it should identify the paper as a biomedical image analysis challenge and indicate the image modality(ies) applied as well as the task category (e.g. classification, segmentation) corresponding to the challenge. The *abstract* should serve as a high-level summary of the challenge purpose, design and results and report the main conclusions. The *keywords* should comprise the main terms characterizing the challenge.

*B. Introduction: Research Context*

The first section should provide the *challenge motivation and objectives* from both a *biomedical* and *technical* point of view. It should summarize the most important related work and clearly outline the expected impact of the challenge compared to previous studies. The task to be solved/ addressed by the challenge should be explicitly stated, and the section should clarify whether the challenge mainly focuses on comparative benchmarking of existing solutions or whether there is a necessity of improving existing solutions.

*C. Methods: Reporting of challenge design*

The challenge design parameters to be reported are classified in four categories related to the topics *challenge organization, mission of the challenge, challenge data sets*, and *assessment method*. The following paragraphs summarize the information that should be provided in the corresponding subsections.

*1) Challenge organization*

This section should include all of the relevant information regarding challenge organization and participation conditions. This information can either be reported in the main document or be provided as supplementary information (e.g. using the form provided in *Suppl 1*). It should include the *challenge name* (including *acronym* (if any)) as well as information on the *organizing team* and the intended challenge *life cycle type*. Note that not every challenge closes after the submission deadline (one-time event). Sometimes it is possible to submit results after the deadline (open call/continuous benchmarking) or the challenge is repeated with some modifications (repeated event). Information on *challenge venue and platform* should include the event (e.g. conference, if any) that the challenge was associated with, the platform that was applied to run the challenge as well as a link to the *challenge website* (if any).

Comprehensive information about *participation policies* should be related to the *interaction level policy* (e.g. only fully-automatic methods allowed), the *training data policy* (indicating which data sets could be used to complement the data sets provided by the challenge (if any)), the *award policy*, which typically refers to challenge prizes and the *results announcements* (e.g. only names of top 3 performing teams will be publicly announced). It should also contain information about the *organizer participation policy*. A policy related to this aspect may be, for example, that members of the organizers' institutes could participate in the challenge but are not eligible for awards and are not listed in the leaderboard. Crucially, annotators of the test data should generally not be allowed to annotate additional training data that is exclusively provided to only one/some of the participating teams (Remark of the authors: Such a case of intentional or unintentional "cheating" has occurred in the past). Finally, details on the *publication policy* should be provided: Do all participating teams automatically qualify as co-authors? Or only the top performing ones (theoretically, this could prevent people from participating with an arbitrary method just for the sake of being an author of a highly cited paper)? Who of the participating teams' members qualifies as an author (e.g. fixed maximum number per team? All team members? First author and supervisor (if any)?)? Can the participating teams publish their results separately? If so: After an embargo?

The section should further contain information on the *submission method*, preferably including a link to the *instructions* that the participating teams received. It should also include information on the procedure for *evaluating the algorithms before the best runs/the final method were submitted* for final performance assessment.

Information on the *challenge schedule* should focus on the date(s) of training, validation (if any) and test data release as well as on the submission of algorithm results on test data, the associated workshop days (if any) and the release date of the results. In some challenges, a post-competition collaborative phase can take place, where e.g. top teams are brought together to further improve on solutions. This should be explicitly mentioned in the schedule.

Crucially, information related to the challenge organization should include information on the *ethics approval* (if applicable) and the *data usage agreement* (indicating who may use the data for which purposes under which conditions). Similarly, information on *code availability* should be provided explicitly relating to both the *organizers'* and the *participating teams'* software. To make *conflicts of interest* transparent, the section should also list the funding/sponsoring associated with the challenge and should explicitly state who had access to the test case(s) labels and when. Finally, the *author contributions* should be explicitly listed in the supplementary material.

*2) Mission of the challenge*

This paragraph should state the biomedical application (*field of application*, e.g. diagnosis, screening, intervention planning) and the *task category* (e.g. segmentation, classification, retrieval, detection) that the participating teams' algorithms were designed for. To refer to the subjects (e.g. patients)/objects (e.g. physical phantoms) from whom/which the image data was acquired, we use the term *cohort*. The paper should explicitly distinguish between the *target cohort*, which refers to the subject(s)/object(s) from whom/which the data would be acquired in the final biomedical application (e.g. healthy subjects from Germany who undergo screening) and the *challenge cohort*, defined as the subject(s)/object(s) from whom/which the challenge data was acquired (e.g. white male healthy subjects from Germany who participated in a voluntary study X). Note that this differentiation is crucial to understand the potential "domain gap'" when transferring challenge results to the actual application. Important differences may be related to various aspects including the subject(s)/object(s) from whom/which the image data are acquired (e.g. cancer patients in real application vs. porcine models in a challenge), the source of image data (e.g. various CT scanners in real application vs. a specific scanner in the challenge) and the characteristics of data (e.g. fraction of malignant cases in real world vs. equal number of malignant and benign cases in the challenge).

To describe the cohorts in detail, the section should also include information on the *imaging modalities*, and additional *context information* (e.g. clinical data) acquired. Most challenges performed to date are based solely on images and corresponding reference annotations (e.g. tumor labels), yet an increasing number of competitions provide further information on the patients, such as general information (age, gender), or laboratory results.

The section should further state the *target entity(ies)* which includes the *data origin*, i.e. the region from which the image data are acquired (e.g. scan of the head, video of the whole operating theater) and the *algorithm target* defined as the structure (e.g. tumor in the brain), object (e.g. robot), subject (e.g. nurse) or component (e.g. tip of a medical instrument) that the participating algorithms focus on.

Finally, it should provide a concise statement of the *assessment aim(s)* (e.g. finding the most sensitive lesion detection algorithm vs. identifying the fastest algorithm that provides a median detection accuracy below a certain threshold). The metric(s) and ranking scheme chosen (parameters 29 and 30 in *assessment method*, Appendix A [22]) should reflect the assessment aims as closely as possible, i.e. optimizing the metrics will ideally optimize the properties of the algorithm that are important according to the assessment aim. Note that it is necessary to make the assessment aim explicit, as it may not be straightforward to find an appropriate metric for certain properties to be optimized.

*3) Challenge data sets*

While the information contained in the challenge mission section should refer to both the target cohort and the challenge cohort, this section is exclusively dedicated to the challenge cohort. It should start with a description on the *data source(s)*. This should include information on specific *acquisition devices* (e.g. the specific type of magnetic resonance imaging (MRI) scanner used), *acquisition protocols* (e.g. the specific MRI imaging protocol applied) as well as *centers/data providing source(s)* and *operators* that were involved in the data acquisition (e.g. a specific robot in a specific university clinic). If the centers involved cannot be mentioned due to requirements for anonymity, this should be made explicit. Information on the operators should focus on the *relevant information in the challenge context*. It may, for example, be irrelevant to state the years of experience of the person acquiring an MRI image according to an established protocol whereas, for data derived from a complex surgical procedure, it may be crucially important to explicitly list the level of expertise of the surgeon.

The section should further provide information on the *training and test case characteristics*. It should begin with stating explicitly *what data encompasses a single case*, i.e. which data are meant to be processed to produce one result that is compared to the corresponding reference result. Information on the cases should further include information on the *number of training/test cases* as well as on *why* a specific proportion of training/test data was chosen, why a certain total number of cases was chosen and why certain *characteristics* were chosen for the training/test set (e.g. class distribution according to real-world distribution vs. equal class distribution).

Information on the *annotation characteristics* should begin with describing the *general approach to training/test case annotation* (e.g. annotation of the test data by a medical expert vs. annotation of the training data via crowdsourcing such as in [28]). It should include the *instructions* given to the annotators prior to the annotation, details on the *subject(s)/algorithm(s) that annotated the cases* (e.g. information on level of expertise such as number of years of professional experience, medically-trained or not) and the *method(s) used to merge multiple annotations* for one case. All information should be provided separately for the training, validation and test cases if necessary.

*Data pre-processing methods* (if any) used to process the raw data should also be well-described and justified. Crucially, potential *sources of errors* related to the *annotation* but also the *data acquisition* should be comprehensively described. Sources of error related to the data acquisition may, for example, refer to calibration errors of the image modality, tracking errors related to pose computation of surgical instruments or errors resulting from substantial motion during image acquisition. Preferably, a quantitative analysis (e.g. using the concept of intra-annotator and inter-annotator variability) should be performed to estimate the magnitude of the different error sources.

*4) Assessment method*

The *metric(s)* to assess a property of an algorithm should be well-explained including the parameters used (if any) and preferably with reference to a paper. The metrics should be justified in the context of the challenge objective (parameter *assessment aim(s)*) and the biomedical application. For example, the Dice similarity coefficient (DSC) is a well-suited metric for assessing segmentation accuracy for large structures, such as organs in images, but is not well-suited for quantifying segmentation accuracy in the case of small pathologies.

If one or multiple rankings were generated for the challenge, the *ranking method(s)* should be specified by describing how metric values are aggregated/used to generate a final ranking (if any). It should also provide information on the rank value in case of tied positions, as well as on methods used to manage submissions with missing results on test cases (*missing data handling*) and to handle any diversity in the level of user interaction when generating the performance ranking (interaction level handling). The section should further make explicit how the ranking chosen matches the assessment aim.

Details for all the statistical methods used to analyze the data should be provided. If results on test cases were entered as missing values, it should be described how these were handled in the *statistical analyses*. Further, details about the assessment of the robustness of the ranking should be provided. If statistical hypothesis tests were used to compare, e.g. participating teams, details about the statistical method should be provided including a description of any method used to assess whether the data met the assumptions required for the particular statistical approach. For all data analysis methods, the software product used should be mentioned. Preferably, the code should also be released along with the paper.

*Further analyses* performed should also be reported in this section. This includes experiments based on combining individual algorithms via ensembling and experiments on inter-algorithm variability, for example.

*D. Results: Reporting of Challenge Outcome*

We suggest subdividing the outcome section into five categories: *Challenge submission, information on selected participating teams, metric values, rankings* and *further analyses*.

At first, information on the *submissions* received should be summarized. This includes the *number of registrations*, the *number of (valid) submissions* (if applicable: in each phase) and the *number of participating teams* that the challenge paper is about (selected participating teams) with *justification* why these participating teams were chosen (e.g. top *n* performing teams; teams with a metric value above a threshold; *m%* top performing teams). Depending on the number of participating teams, *information on selected participating teams* can be provided in the main document or in the appendix. Information on those teams referred to in the results section should include a *team identifier* (as name of the team) as well as a *description of the method*. The latter can for example be a brief textual summary plus a link to a document that provides detailed information not only on the basic method but also on the specific parameters /optimizations performed for the challenge. Ideally, this reference document should also provide information on complexity analysis with respect to time and memory consumption, hardware/OS requirements and reference to the source code.

Depending on the number of test cases and participating teams, raw metric values (i.e. metric values for each test case) and/or aggregated *metric values* should be provided for all participating teams/the selected teams. Parts of these results may be moved to the appendix.

The *ranking(s)* (if any) should be reported including the results on robustness analyses (e.g. bootstrapping results [29]) and other *statistical analyses*. Again, depending on the number of participating teams, the paper may refer to only the top performing teams (referred to as selected participating teams above). Depending on the number of participating teams, full (if necessary partially anonymized) ranking(s) should be provided in the main document, as supplementary material or in another citable document.

The results of *further analyses* performed (if any) should also be reported in this section. This includes analyses of common problems/biases of the methods.

*E. Discussion: Putting the Results into Context*

The final section should provide a concise *summary* of the challenge outcome and discuss the findings of the challenge thoroughly in the context of the state of the art. It should clearly distinguish between the *technical* and *biomedical impact*. Current performance of the best methods should be *discussed* and conclusions drawn about *whether the task is already solved* in a satisfactory way (e.g. the remaining errors are comparable to inter-annotator variability). Furthermore, an *analysis of individual cases*, in which the majority of algorithms performed poorly (if any), should be included. Also, *advantages and disadvantages of the participating methods* should be discussed. In this context, it should be made explicit whether an algorithm with clearly superior performance could be identified or if more than one algorithm is well-suited for the specific task. Furthermore, *limitations of the challenge* should be made explicit (design *and* execution). Finally, concrete *recommendations for future work* should be provided and a *conclusion* drawn.

## III. DISCUSSION

As a first step to address the discrepancy between the impact of biomedical image analysis challenges and the quality (control), the BIAS initiative aims to improve the transparency of the *reporting*. This article describes how the BIAS statement was developed and presents a checklist which authors of biomedical image analysis challenges are encouraged to include in their submission when giving a challenge paper into review. By making relevant information explicit, the checklist has the purpose to standardize and facilitate the reviewing/editorial process and raise interpretability and reproducibility of challenge results.

The checklist generated in the scope of this article relies heavily on the challenge parameter list published in our previous work [18]. In the meantime, this parameter list has been instantiated with more than 500 tasks from more than 150 challenges, both retrospectively [18] and prospectively in the scope of the structured challenge submission system used for MICCAI 2018 and 2019. According to our experience, the (updated) list presented in this work should be appropriate to capture the relevant information on current challenge design and organization. It is worth noting, however, that an update of the checklist may be required at a later point in time. For example, ongoing research is investigating the generation of *probabilistic output* for a whole range of algorithm categories [30]; rather than providing a single contour as result for a segmentation task. For instance, such methods produce a whole range of *plausible* solutions via sampling. It is currently unknown how such output may be efficiently handled in the design of future challenges.

An increasingly relevant problem is that it typically remains unknown which specific feature of an algorithm actually makes it better than competing algorithms [18]. For example, many researchers are convinced that the method for data augmentation often has a much bigger influence on the performance of a deep learning algorithm than the network architecture itself. For this reason, a structured description (e.g. using ontologies) not only of the challenge but also of the participating algorithms may be desirable. Due to the lack of common software frameworks and terminology, however, this is not trivial to implement at this stage [18].

It is worth mentioning that our guideline has explicitly been developed for reporting the design and results for *one task* of a challenge. If a challenge includes multiple tasks, the results should preferably be reported in separate publications. If this is not desirable (i.e. a single paper refers to multiple tasks with substantial overlap between tasks such as tasks sharing the same data sets), a separate checklist for each task should be generated. Alternatively, a single checklist may be provided in which some items are common to all tasks and other items contain separate parts for each task.

It should also be noted that challenges could in theory focus on collaboration rather than competition. In such collaborative challenges, the participating teams would work jointly on a dedicated problem, and the focus would be on solving a problem together rather than benchmarking different methods. We have not explicitly addressed such collaborative challenges with the checklist.

We believe that the work invested to improve challenge reporting could also be valuable in guiding challenge design. For this reason, we have converted the reviewer checklist into a document that can be used to comprehensively report the envisioned design of a challenge and could thus be used to review a challenge before it is organized (*Suppl 2*). Based on this document, the MICCAI society and MICCAI 2020 organizing team decided to introduce the concept of challenge registration. Similar to how clinical trials have to be registered before they are started, the complete design of accepted MICCAI challenges had to be put online before challenge execution. This was achieved with Zenodo [1], a general-purpose open-access repository that allows researchers to deposit data sets, software, and other research-related items. Such stored items are citable, because a persistent digital object identifier (DOI) is generated for each submission. As Zenodo also supports version control, changes to a challenge design (e.g. to the metrics or ranking schemes applied) can be made transparent. These changes must be communicated to the MICCAI society and well-justified. To date (June 2020), 8 out of the 28 challenges committed changes to the designs, originally uploaded in April 2020. Most of them were changes to the schedule, which can be attributed to the COVID-19 outbreak. We believe that the transparency and quality control that comes along with challenge registration is a big step towards higher quality of biomedical challenges.

Challenges are becoming increasingly important in various fields, ranging from protein structure, to systems biology, text mining, and genomics, thanks to initiatives such as CASP (Critical Assessment of Techniques for Protein Structure Prediction) [31], BioCreative (Critical Assessment of Information Extraction in Biology [32]), DREAM (Dialogue for Reverse Engineering Assessment and Methods [26]), and CAGI (Critical Assessment of Genome Interpretation [33]). The checklist and the challenge design document could be adapted to these research areas and thus contribute substantially to better scientific practice related to challenges in general.

In conclusion, this document is the first to provide a guideline for the reporting of a biomedical image analysis challenge regardless of field of application, image modality or algorithm category assessed. We hope that the checklist provided will help editors of journals in the field of biomedical image analysis and beyond to establish a rigorous review process with the mid-term goal of increasing interpretability and reproducibility of results and raising the quality of challenge design in general.

COMPETING INTERESTS

Anne Martel is a co-founder, CSO of Pathcore, Toronto, Canada. The remaining authors declare no competing interests.

ACKNOWLEDGMENTS

The project was conducted in the scope of the Helmholtz Imaging Platform (HIP) funded by the Helmholtz Association of German Research Centres. We acknowledge support from the European Research Council (ERC) under the New Horizon Framework Programme grant agreement ERC-2015-StG-37960 (ERC starting grant COMBIOSCOPY) as well as the Natural Science and Engineering Research Council (NSERC) (RGPIN-2016-06283), the Canadian Cancer Society (#705772), the Ministry of Education, Youth and Sports of the Czech Republic (Projects LTC17016 and CZ.02.1.01/0.0/0.0/16\013/0001775), the National Science Foundation 1452485 and the National Institutes of Health R01-EB017230-01A1.

REFERENCES

[1] Andre Esteva et al. A guide to deep learning in healthcare. *Nat. Med.*, 25(1):24, 2019. doi: 10.1038/s41591-018-0316-z.

[2] Eric J Topol. High-performance medicine: the convergence of human and artificial intelligence. *Nat. Med.*, 25(1):44, 2019. doi: 10.1038/s41591-018-0300-7.

[3] Nicholas Ayache and James Duncan. 20[th] anniversary of the medical image analysis journal (MedIA). *Med. ImageAnal.*, 33:1–3, 2016. doi: 10.1016/j.media.2016.07.004.

[4] Ahmed Hosny et al. Artificial intelligence in radiology. *Nat. Rev.Cancer*, 18:500, 2018. doi: 10.1038/s41568-018-0016-5.

[5] Geert Litjens et al. A survey on deep learning in medical image analysis. *Med. ImageAnal.*, 42:60–88, 2017. doi: 10.1016/j.media.2017.07.005.

[6] Kangkana Bora et al. Automated classification of pap smear images to detect cervical dysplasia. *Comput. Methods Programs Biomed.*, 138:31–47, 2017. doi: 10.1016/j.cmpb.2016.10.001.

[7] Dmitrii Bychko et al. Deep learning based tissue analysis predicts outcome in colorectal cancer. *Sci. Rep.*, 8(1):3395, 2018.doi: 10.1038/s41598-018-21758-3.

[8] Kun-Hsing Yu et al. Predicting non-small cell lung cancer prognosis by fully automated microscopic pathology image features. *Nat. Commun.*, 7:12474, 2016.doi: 10.1038/ncomms12474.

[9] Marcia L Meldrum. A brief history of the randomized controlled trial: From oranges and lemons to the gold standard. *Hematol. Oncol. Clin. North Am.*, 14(4):745–760, 2000.doi: 10.1016/S0889- 8588(05)70309-9.

[10] Michal Kozubek. Challenges and benchmarks in bioimage analysis. In *Focus Bio Image Inform.*, pages 231–262. Springer, 2016. doi: 10.1007/978-3-319-28549-8_9.

[11] Nicolas Chenouard et al.Objective comparison of particle tracking methods. *Nat. Methods*, 11(3):281–289, 2014. doi: 10.1038/nmeth.2808.

[12] Klaus H Maier-Hein et al. The challenge of mapping the human connectome based on diffusion tractography. *Nat. Commun.*, 8(1):1349,2017.doi: 10.1038/s41467-017-01285-x.

[13] Bjoern H Menze et al. The multimodal brain tumor image segmentation benchmark (BRATS). *IEEE Trans.Med.Imaging*, 34(10):1993–2024, 2015.doi: 10.1109/TMI.2014.2377694.

[14] Daniel Sage et al.Quantitative evaluation of software packages for single-molecule localization microscopy. *Nat. Methods*, 12(8):717–724, 2015. doi: 10.1038/nmeth.3442.

[15] Arnaud Arindra Adiyoso Setio et al. Validation,comparison,and combination of algorithms for automatic detection of pulmonary nodules in computed tomography images: the LUNA16 challenge. *Med. Image Anal.*, 42:1–13,2017. doi: 10.1016/j.media.2017.06.015.

[16] Guoyan Zheng, et al. *Med. Image Anal.*, 35:327–344, 2017.doi: 10.1016/j.media.2016.08.005.

[17] Kaggle. Your home for data science. www.kaggle.com/, 2010. Accessed: 2019-02-12.

---

[1] https://zenodo.org/

[18] Lena Maier-Hein et al. Why rankings of biomedical image analysis competitions should be interpreted with care. *Nat. Commun.*, 9(1):5217, 2018. doi: 10.1038/s41467-019-08563-w.

[19] The EQUATOR Network. The EQUATOR network – Enhancing the QUAlity and Transparency Of health Research. http://www.equator-network.org, 2008. Accessed: 2019-09-12.

[20] David Moher, Kenneth F Schulz, and Douglas G Altman. The consort statement: revised recommendations for improving the quality of reports of parallel group randomized trials. *BMC medicalresearchmethodology*, 1(1):2, 2001.

[21] Kenneth F Schulz, Douglas G Altman, and David Moher. Consort2010 statement: updated guidelines for reporting parallel group randomised trials. *BMC medicine*, 8(1):18, 2010.

[22] Lena Maier-Hein et al. Biomedical Image Analysis Challenges (BIAS) Reporting Guideline. *Zenodo*, 2020. doi: 10.5281/zenodo.4008953.

[23] Pierre Jannin, Christophe Grova, and Calvin R Maurer. Model for defining and reporting reference-based validation protocols in medical image processing. *Int. J.CARS*, 1(2):63–73, 2006. doi: 10.1007/s11548-006-0044-6.

[24] Justin Guinney and Julio Saez-Rodriguez. Alternative models for sharing confidential biomedical data. *Nat. Biotechnol.*, 36(5):391, 2018. doi: 10.1038/nbt.4128.

[25] DREAM. Dream challenges. dreamchallenges.org/, 2006. Accessed: 2019-04-01.

[26] Julio Saez-Rodriguez et al. Crowdsourcing biomedical research: leveraging communities as innovation engines. *Nat. Rev.Genet.*, 17(8):470–486, 2016. doi: 10.1038/nrg.2016.69.

[27] crowdAI. crowdAI. www.crowdai.org/, 2018. Accessed: 2019-04-01.

[28] Lena Maier-Hein et al. Crowdsourcing for reference correspondence generation in endoscopic images. In *Int. Conf. Med. ImageComput.Comp.Assis.Interv.*, pages 349–356. Springer, 2014.

doi: 10.1007/978-3-319-10470-6_44.

[29] Wiesenfarth et al. Methods and open-source toolkit for analyzing and visualizing challenge results. 2019. *arXiv preprint arXiv:1910.05121*.

[30] Simon Kohl et al. A probabilistic U-net for segmentation of ambiguous images. In *Adv. Neural Inf. Process. Syst.(NIPS2018)*, volume 31, pages 6965–6975. 2018. ArXiv version at http://arxiv.org/abs/1806.05034.

[31] John Moult et al. Critical assessment of methods of protein structure prediction(CASP)–RoundXII. *Proteins*, 86:7–15, 2017.

doi: 10.1002/prot.25415.

[32] Rezarta Islamaj Dogan, et al. Overview of the Bio Creative VI precision medicine track: mining protein interactions and mutations for precision medicine. *Database (Oxford)*, 2019, 2019.

doi: 10.1038/10.1093/database/bay147.

[33] Roxana Daneshjou et al. Working toward precision medicine: Predicting phenotypes from exomes in the critical assessment of genome interpretation (CAGI) challenges. *Hum. Mut.*, 38(9):1182–1192, 2017. doi: 10.1002/humu.23280.

APPENDIX

# Appendix A: BIAS Reporting Guideline

| Section/ Topic | Parameter name | Item No | Checklist Item | Reported on page No |
|---|---|---|---|---|
| **TITLE, ABSTRACT, KEYWORDS** | Title | 1 | Use the title to convey the essential information on the **challenge mission**.<br>The title should ...<br>● … identify the paper as biomedical image analysis challenge.<br>● … indicate the image modality(ies) applied with a commonly used term in the title.<br>● … indicate the task and/or task category (e.g. classification, segmentation; see parameter 18) with a commonly used term in the title.<br>● … (optionally) include information on the biomedical target application.<br>● … (optionally) include the year for repeated challenges with fixed cycle. | |
| | Abstract | 2 | Provide a **summary** of the challenge purpose, design and results and report the main conclusion(s). | |
| | Keywords | 3 | List the primary **keywords** that characterize the challenge. | |
| **INTRO-DUCTION** | Challenge motivation and objective | 4a | Provide a general introduction to the topic from a **biomedical point of view**. This should include the envisioned biomedical impact (short-term and/or long-term). | |
| | | 4b | Provide a general introduction to the topic from a **technical point of view**. This should include an overview of the state of the art along the envisioned technical/methodological impact. | |
| | | 4c | Based on the biomedical and technical motivation, provide a concise statement of the **primary challenge objective**. This should include a **statement of the task**. | |
| **METHODS Challenge organi-zation** | Challenge name | 5a | Provide a **representative name** of the challenge.<br>*Example:* MICCAI Endoscopic Vision Challenge 2015 | |
| | | 5b | Provide the **acronym** of the challenge (if any).<br>*Example:* EndoVis15 | |
| | Organizing team | 6 | Provide information on the **organizing team** (names and affiliations). | |

| | | |
|---|---|---|
| **Life cycle type** | 7 | Define the intended **submission cycle** of the challenge. Include information on whether/how the challenge has been/will be continued after the present study.<br>*Examples:*<br>● One-time event with fixed submission deadline<br>● Open call<br>● Repeated event with annual fixed submission deadline |
| **Challenge venue and platform** | 8a | Report the **event** (e.g. conference) that was **associated** with the challenge (if any). |
| | 8b | Report the **platform** (e.g. grand-challenge.org) used to run the challenge. |
| | 8c | Provide the **URL** for the challenge website (if any). |
| **Participation policies** | 9a | Define the **allowed user interaction** of the algorithms assessed (e.g. only (semi-) automatic methods allowed). |
| | 9b | Define the policy on the **usage of training data**. The data used to train algorithms may, for example, have been restricted to the data provided by the challenge or to publicly available data including (open) pre-trained nets. |
| | 9c | Define the **participation policy for members of the organizers' institutes**. For example, members of the organizers' institutes could participate in the challenge but were not eligible for awards. |
| | 9d | Define the **award policy**. In particular, provide details with respect to challenge prizes. |
| | 9e | Define the policy for **results announcement**.<br>*Examples:*<br>● Top three performing methods were announced publicly.<br>● Participating teams could choose whether the performance results will be made public. |
| | 9f | Define the **publication policy**. In particular, provide details on …<br>● … who of the participating teams/the participating teams' members qualified as author<br>● … whether the participating teams could publish their own results separately, and (if so) |
| | | ● … whether an embargo time was defined (so that challenge organizers can publish a challenge paper first). |
| **Submission method** | 10a | Describe the method used for result submission. If available, provide a link to the **submission instructions**.<br>*Examples:*<br>● Docker container on the Synapse platform. Link to submission instructions: <URL><br>● Algorithm output was sent to organizers via e-mail. Submission instructions were sent by e-mail. |
| | 10b | Provide information on the possibility for participating teams to **evaluate** their **algorithms before submitting** final results. For example, many challenges allow submission of multiple results, and only the last run is officially counted to compute challenge results. |
| **Challenge schedule** | 11 | Provide a **timetable** for the challenge. Preferably, this should include<br>● the release date(s) of the training cases (if any)<br>● the registration date/period<br>● the release date(s) of the test cases and validation cases (if any)<br>● the submission date(s)<br>● associated workshop days (if any)<br>● the release date(s) of the results |
| **Ethics approval** | 12 | Indicate whether **ethics approval** was necessary for the data. If yes, provide details on the ethics approval, preferably institutional review board, location, date and number of the ethics approval (if applicable). Add the URL or a **reference to the document** of the ethics approval (if available). |
| **Data usage agreement** | 13 | Clarify how the data can be used and distributed by the teams that participate in the challenge and by others. This should include the explicit **listing of the license** applied.<br>*Examples:*<br>● CC BY (Attribution)<br>● CC BY-SA (Attribution-ShareAlike)<br>● CC BY-ND (Attribution-NoDerivs)<br>● CC BY-NC (Attribution-NonCommercial) |

| | | | |
|---|---|---|---|
| | | | - CC BY-NC-SA (Attribution-NonCommercial-ShareAlike)<br>- CC BY-NC-ND (Attribution-NonCommercial-NoDerivs) |
| | **Code availability** | 14a | Provide information on the **accessibility of the organizers' evaluation software** (e.g. code to produce rankings). Preferably, provide a link to the code and add information on the supported platforms. |
| | | 14b | In an analogous manner, provide information on the **accessibility of the participating teams' code.** |
| | **Conflicts of interest** | 15 | Provide information related to conflicts of interest. In particular provide information related to **sponsoring/funding** of the challenge. Also, state explicitly who had **access to the test case labels** and when. |
| | **Author contributions** | 16 | List the **contributions** of all authors to the paper (preferably in the appendix). |
| **METHODS Mission of the challenge** | **Field(s) of application** | 17 | State the **main field(s) of application** that the participating algorithms target.<br>*Examples:*<br>- Diagnosis<br>- Education<br>- Intervention assistance<br>- Intervention follow-up<br>- Intervention planning<br>- Prognosis<br>- Research<br>- Screening<br>- Training<br>- Cross-phase |
| | **Task category(ies)** | 18 | State the **task category(ies)**.<br>*Examples:*<br>- Classification<br>- Detection<br>- Localization<br>- Modeling<br>- Prediction<br>- Reconstruction<br>- Registration<br>- Retrieval |

| | | |
|---|---|---|
| | | ● Segmentation |
| | | ● Tracking |
| **Cohorts** | | We distinguish between the *target cohort* and the *challenge cohort*. For example, a challenge could be designed around the task of medical instrument tracking in robotic kidney surgery. While the challenge could be based on *ex vivo* data obtained from a laparoscopic training environment with porcine organs (challenge cohort), the final biomedical application (i.e. robotic kidney surgery) would be targeted on real patients with certain characteristics defined by inclusion criteria such as restrictions regarding gender or age (target cohort). |
| | 19a | Describe the **target cohort**, i.e. the subjects/objects from whom/which the data would be acquired in the final biomedical application. |
| | 19b | Describe the **challenge cohort**, i.e. the subject(s)/object(s) from whom/which the challenge data was acquired. |
| **Imaging modality(ies)** | 20 | Specify the **imaging technique(s)** applied in the challenge. |
| **Context information** | | Provide additional **information given along with the images**. The information may correspond ... |
| | 21a | … directly to the **image data** (e.g. tumor volume). If necessary, differentiate between target and challenge cohort. |
| | 21b | … to the **patient** in general (e.g. gender, medical history). If necessary, differentiate between target and challenge cohort. |
| | 21c | … to the **acquisition process** (e.g. medical device data during endoscopic surgery, calibration data for an image modality). If necessary, differentiate between target and challenge cohort. |
| **Target entity(ies)** | 22a | Describe the **data origin**, i.e. the region(s)/part(s) of subject(s)/object(s) from whom/which the image data would be acquired in the final biomedical application (e.g. brain shown in computed tomography (CT) data, abdomen shown in laparoscopic video data, operating room shown in video data, thorax shown in fluoroscopy video). If necessary, differentiate between target and challenge cohort. |
| | 22b | Describe the **algorithm target**, i.e. the structure(s)/subject(s)/object(s)/component(s) that the participating algorithms have been designed to focus on (e.g. tumor in the brain, tip of a medical instrument, nurse in an operating theater, catheter in a fluoroscopy scan). If necessary, differentiate between target and challenge cohort. |
| **Assessment aim(s)** | 23 | Identify the **property(ies) of the algorithms to be optimized** to perform well in the challenge. If multiple properties were |

| | | | |
|---|---|---|---|
| | | | assessed, prioritize them (if appropriate). The properties should then be reflected in the metrics applied (parameter 29), and the priorities should be reflected in the ranking when combining multiple metrics that assess different properties.<br>● *Example 1:* Find liver segmentation algorithm for CT images that processes CT images of a certain size in less than a minute on a certain hardware with an error that reflects inter-rater variability of experts.<br>● *Example 2:* Find lung tumor detection algorithm with high sensitivity and specificity for mammography images.<br>Corresponding metrics are listed below (parameter 29). |
| **METHODS Challenge data sets** | **Data source(s)** | 24a | Specify the **device(s)** used to acquire the challenge data. This includes details on the device(s) used to acquire the imaging data (e.g. manufacturer) as well as information on additional devices used for performance assessment (e.g. tracking system used in a surgical setting). |
| | | 24b | Describe relevant details on the imaging process/**data acquisition** for each acquisition device (e.g. image acquisition protocol(s)). |
| | | 24c | Specify the **center(s)/institute(s)** in which the data was acquired and/or the **data providing platform/source** (e.g. previous challenge). If this information is not provided (e.g. for anonymization reasons), specify why. |
| | | 24d | Describe relevant **characteristics** (e.g. level of expertise) **of the subjects** (e.g. surgeon)/objects (e.g. robot) involved in the data acquisition process (if any). |
| | **Training and test case characteristics** | 25a | State what is meant by one **case** in this challenge. A case encompasses all data that is processed to produce one result that is then compared to the corresponding reference result (i.e. the desired algorithm output).<br>*Examples:*<br>● Training and test cases both represented a CT image of a human brain. Training cases had a weak annotation (tumor present or not and tumor volume (if any)) while the test cases were annotated with the tumor contour (if any). |

| | | |
|---|---|---|
| | | • A case refers to all information that is available for one particular patient in a specific study. This information always includes the image information as specified in *data source(s)* (parameter 24) and may include context information (parameter 21). Both training and test cases were annotated with survival (binary) 5 years after (first) image was taken. |
| | 25b | State the **total number** of cases as well as the number of training, validation and test cases separately. |
| | 25c | Explain **why a total number** of cases and **the specific proportion** of training, validation and test cases was chosen. |
| | 25d | Mention **further important characteristics** of the training, validation and test cases (e.g. class distribution in classification tasks chosen according to real-world distribution vs. equal class distribution) and justify the choice. |
| **Annotation characteristics** | 26a | Describe the **method for determining the reference annotation**, i.e. the desired algorithm output. Provide the information separately for the training, validation and test cases if necessary. Possible methods include *manual image annotation*, *in silico ground truth generation* and *annotation by automatic methods*. If human annotation was involved, state the **number of annotators**. |
| | 26b | Provide the **instructions given to the annotators** (if any) prior to the annotation. This may include description of a training phase with the software. Provide the information separately for the training, validation and test cases if necessary. Preferably, provide a link to the **annotation protocol**. |
| | 26c | Provide **details on the subject(s)/algorithm(s) that annotated** the cases (e.g. information on **level of expertise** such as number of years of professional experience, medically-trained or not). Provide the information separately for the training, validation and test cases if necessary. |
| | 26d | Describe the **method(s) used to merge multiple annotations** for one case (if any). Provide the information separately for the training, validation and test cases if necessary. |
| **Data pre-processing method(s)** | 27 | Describe the **method(s) used for pre-processing** the raw training data before it is provided to the participating teams. Provide the |

| | | | |
|---|---|---|---|
| | | | information separately for the training, validation and test cases if necessary. |
| | Sources of error | 28a | Describe the most relevant **possible error sources related to the image annotation**. If possible, **estimate the magnitude** (range) of these errors, using inter-and intra-annotator variability, for example. Provide the information separately for the training, validation and test cases, if necessary. |
| | | 28b | In an analogous manner, describe and quantify **other relevant sources of error**. |
| METHODS Assess-ment methods | Metric(s) | 29a | Define the **metric(s) to assess a property of an algorithm**. These metrics should reflect the desired algorithm properties described in *assessment aim(s)* (parameter 21). State which metric(s) were used to compute the ranking(s) (if any).<br>● *Example 1:* Dice Similarity Coefficient (DSC) and run-time<br>● *Example 2:* Area under curve (AUC) |
| | | 29b | **Justify why** the metric(s) was/were chosen, preferably with reference to the biomedical application. |
| | Ranking method(s) | 30a | Describe the **method used to compute a performance rank** for all submitted algorithms based on the generated metric results on the test cases. Typically the text will describe how results obtained per case and metric are aggregated to arrive at a final score/ranking. |
| | | 30b | Describe the method(s) used to manage **submissions with missing results** on test cases. |
| | | 30c | **Justify why** the described ranking scheme(s) was/were used. |
| | Statistical analyses | 31a | Provide **details for all statistical methods** used in the scope of the challenge analysis. This may include<br>● description of the **missing data handling**,<br>● details about the assessment of **variability of rankings**,<br>● description of any method used to assess **whether the data met the assumptions**, required for the particular statistical approach, or<br>● indication of any **software product** that was used for data analysis. |
| | | 31b | **Justify why** the described statistical method(s) was/were used. |
| RESULTS | Challenge submissions | | Provide **summarizing information** on ... |
| | | 32a | … the **number of registrations**. |
| Challenge outcome | | 32b | … the **number of participating teams** that provided valid submissions (if applicable in each phase). |
| | | 32c | … the **number of participating teams** that the **paper** refers to (with **justification**). |
| | Information on selected participating teams | | Provide the following information for the participating teams that are included in the paper: |
| | | 33a | **Team identifier**. |
| | | 33b | A **method description** including parameter instantiation and/or a reference/URL to a document containing this information. |
| | Metric values | 34 | Provide raw and/or **aggregated metric values** (including measure of variability) for all participating teams and each metric (if applicable) as well as the **numbers of test set submissions** (the last one was used to compute metric(s)) for each participating team. |
| | Ranking(s) | 35a | Report the **ranking(s)** (if any) including the number of test set submissions for each participating team. |
| | | 35b | Provide the results of the **statistical analyses**. |
| | Further Analyses | 36 | Present results of further analyses (if applicable), e.g. related to<ul><li>**combining algorithms** via ensembling,</li><li>**inter-algorithm variability**,</li><li>**common problems/biases** of the submitted methods, or</li><li>**ranking variability**.</li></ul> |
| DISCUS-SION | Summary | 37 | Summarize the **main results** of the challenge. |
| | Impact | 38a | Describe the (expected) **biomedical impact** of the challenge in the context of the state of the art with reference to the challenge motivation (parameter 4a). |
| | | 38b | Describe the (expected) **technical impact** of the challenge in the context of the state of the art with reference to the challenge motivation (parameter 4b). |
| | Discussion of challenge results | 39a | Provide a detailed discussion and conclusion whether the **task is now solved** in a satisfactory way (e.g. the remaining errors are comparable to inter-annotator variability). |
| | | 39b | Provide a detailed **analysis of individual cases**, in which the majority of algorithms performed poorly (if any). |
| | | 39c | Provide a **discussion on advantages and disadvantages of the submitted methods**. Include time and memory consumption comparison if time and memory were not among the metrics. |

| | | |
|---|---|---|
| **Limitations of the challenge** | **40** | Discuss **limitations** related to the challenge design and execution. |
| **Future work** | **41** | Provide **recommendations for future work** and maintenance plans for the challenge and its website (if any). |
| **Conclusions** | **42** | Provide a **concise conclusion** based on the results of the study. |

# Appendix B: Glossary

| Term | Explanation |
| --- | --- |
| **Biomedical image analysis challenge** | Open competition on a specific scientific problem in the field of biomedical image analysis. |
| **Case** | Data set for which the algorithm(s) of interest produce one result in either the training phase (if any) or the test phase. It must include one or multiple images of a biomedical imaging modality (e.g. a computed tomography (CT) and a magnetic resonance imaging (MRI) image of the same structure) and comprises a reference annotation (public for the training phase and secret for the test phase). |
| **Cohort** | Term used to refer to the subjects (e.g. patients) or objects (e.g. physical phantoms) from which the image data was acquired. The paper should explicitly distinguish between the *target cohort*, which refers to the subjects/objects from whom/which the data would be acquired in the final medical/biological application (e.g. healthy subjects from Germany that undergo screening) and the *challenge cohort*, defined as the subject(s)/object(s) from whom/which the challenge data was acquired (e.g. white male healthy subjects from Germany that participated in a voluntary study X). Note that this differentiation is crucial to understand the potential "domain gap" when transferring challenge results to the actual application. |
| **DREAM Challenges** | Dialogue for Reverse Engineering Assessments and Methods Challenges. Non-profit, collaborative community effort consisting of contributors from across the research spectrum, with a focus on systems biology. |
| **Dry run phase** | Challenge phase in which the challenge organizers themselves work with the data to determine the level of difficulty of the task(s), for example by building simple models to solve the challenge. |
| **Final biomedical application** | Biomedical application that algorithms participating in a challenge are designed for. |
| **Leaderboard** | Ranking of the challenges' participating teams or one of its tasks that is generated or updated by the challenge organizers. |
| **Metric** | A measure (not necessarily metric in the strict mathematical sense) used to compute the performance of a given algorithm for a given case, typically based on the reference annotation. |
| **Ranking scheme** | Algorithm according to which a ranking of the participating teams is produced based on the metric values for the test cases of the challenge. |
| **Reference annotation** | Desired result (available for each test case and typically also for a significant number of training cases) to which the computed results of the participating teams are compared to. The term reference annotation (alternatively: gold standard if produced manually or silver standard if produced by merging multiple computer- |
| | generated results) is preferred over *ground truth* because reference annotations are typically only approximations of the (forever unknown) truth (except for the case of synthetic data). |
| **Sage/Synapse platform** | Non-profit biomedical research organization dedicated to developing predictors of disease and accelerating health research through the creation of open systems, incentives, and standards. |
| **Task** | Subproblem to be solved in the scope of a challenge for which a dedicated ranking/leaderboard is provided (if any). The assessment method (e.g. metric(s) applied) may vary across different tasks of a challenge. |
| **Test phase** | Challenge phase in which participants either upload their algorithm, or they get access to the test cases without the reference annotations and submit the results of their algorithm on the test cases to the challenge organizers. |
| **Training phase** | Phase at the beginning of the challenge in which the challenge organizers provide access to training cases with corresponding reference annotations. |
| **Validation phase** | Challenge phase between training and test phase in which initial rankings (*leaderboards*) are generated to show participants how well their methods generalize. Sometimes also referred to as *leaderboard phase*. |

# Suppl 1: Form for summarizing information on challenge organization

**Challenge name**
a) Provide a **representative name** of the challenge.

*Example:* MICCAI Endoscopic Vision Challenge 2015

b) Provide the **acronym** of the challenge (if any).

*Example:* EndoVis15

**Organizing team**
Provide information on the **organizing team** (names and affiliations).

**Life cycle type**
Define the intended **submission cycle** of the challenge. Include information on whether/how the challenge has been/will be continued after the present study.

*Examples:*

- One-time event with fixed submission deadline
- Open call
- Repeated event with annual fixed submission deadline

**Challenge venue and platform**
a) Report the **event** (e.g. conference) that was **associated** with the challenge (if any).

b) Report the **platform** (e.g. grand-challenge.org) used to run the challenge.

c) Provide the **URL** for the challenge website (if any).

**Participation policies**
a) Define the **allowed user interaction** of the algorithms assessed (e.g. only (semi-)automatic methods allowed).

b) Define the policy on the **usage of training data**. The data used to train algorithms may, for example, have been restricted to the data provided by the challenge or to publicly available data including (open) pre-trained nets.

c) Define the **participation policy for members of the organizers' institutes**. For example, members of the organizers' institutes could participate in the challenge but were not eligible for awards.

d) Define the **award policy**. In particular, provide details with respect to challenge prizes.

e) Define the policy for **results announcement**.

*Examples:*

- Top three performing methods were announced publicly.
- Participating teams could choose whether the performance results will be made public.

f) Define the **publication policy**. In particular, provide details on …

- … who of the participating teams/the participating teams' members qualified as author
- … whether the participating teams could publish their own results separately, and (if so)
- … whether an embargo time was defined (so that challenge organizers can publish a challenge paper first).

**Submission method**

a) Describe the method used for result submission. If available, provide a link to the **submission instructions**.

b) Provide information on the possibility for participating teams to **evaluate** their **algorithms before submitting** final results. For example, many challenges allow submission of multiple results, and only the last run is officially counted to compute challenge results.

### Challenge schedule
Provide a **timetable** for the challenge. Preferably, this should include

- the release date(s) of the training cases (if any)
- the registration date/period
- the release date(s) of the test cases and validation cases (if any)
- the submission date(s)
- associated workshop days (if any)
- the release date(s) of the results

### Ethics approval
Indicate whether **ethics approval** was necessary for the data. If yes, provide details on the ethics approval, preferably institutional review board, location, date and number of the ethics approval (if applicable). Add the URL or a **reference to the document** of the ethics approval (if available).

### Data usage agreement
Clarify how the data can be used and distributed by the teams that participate in the challenge and by others. This should include the explicit **listing of the license** applied.

*Examples:*

- CC BY (Attribution)
- CC BY-SA (Attribution-ShareAlike)
- CC BY-ND (Attribution-NoDerivs)
- CC BY-NC (Attribution-NonCommercial)
- CC BY-NC-SA (Attribution-NonCommercial-ShareAlike)
- CC BY-NC-ND (Attribution-NonCommercial-NoDerivs)

### Code availability
a) Provide information on the **accessibility of the organizers' evaluation software** (e.g. code to produce rankings). Preferably, provide a link to the code and add information on the supported platforms.

b) In an analogous manner, provide information on the **accessibility of the participating teams' code.**

## Conflicts of interest

Provide information related to conflicts of interest. In particular provide information related to **sponsoring/ funding** of the challenge. Also, state explicitly who had **access to the test case labels** and when.

## Author contributions

List the **contributions** of all authors to the paper (preferably in the appendix).

# Suppl 2: Structured description of a challenge design

## SUMMARY

**Item 1: Title**
a) Use the title to convey the essential information on the **challenge mission**.

b) Preferable, provide a short **acronym** of the challenge (if any).

**Item 2: Abstract**
Provide a **summary** of the challenge purpose. This should include a general introduction in the topic from both a biomedical as well as from a technical point of view and clearly state the envisioned technical and/or biomedical impact of the challenge.

**Item 3: Keywords**
List the primary **keywords** that characterize the challenge.

## CHALLENGE ORGANIZATION

**Item 4: Organizers**
a) Provide information on the **organizing team** (names and affiliations).

b) Provide information on the **primary contact person**.

**Item 5: Lifecycle type**
Define the intended **submission cycle** of the challenge. Include information on whether/how the challenge will be continued after the challenge has taken place.
*Examples:*
- One-time event with fixed submission deadline
- Open call
- Repeated event with annual fixed submission deadline

**Item 6: Challenge venue and platform**
a) Report the **event** (e.g. conference) that is **associated** with the challenge (if any).

b) Report the **platform** (e.g. grand-challenge.org) used to run the challenge.

c) Provide the **URL** for the challenge website (if any).

**Item 7: Participation policies**
a) Define the **allowed user interaction** of the algorithms assessed (e.g. only (semi-) automatic methods allowed).

b) Define the policy on the **usage of training data**. The data used to train algorithms may, for example, be restricted to the data provided by the challenge or to publicly available data including (open) pre-trained nets.

c) Define the **participation policy for members of the organizers' institutes**. For example, members of the organizers' institutes may participate in the challenge but are not eligible for awards.

d) Define the **award policy**. In particular, provide details with respect to challenge prizes.

e) Define the policy for **result announcement**.
   *Examples:*
   - Top three performing methods will be announced publicly.
   - Participating teams can choose whether the performance results will be made public.

f) Define the **publication policy**. In particular, provide details on ...
   - … who of the participating teams/the participating teams' members qualifies as author
   - … whether the participating teams may publish their own results separately, and (if so)
   - … whether an embargo time is defined (so that challenge organizers can publish a challenge paper first).

**Suppl 2:** Structured description of a challenge design

*BIAS: Transparent reporting of biomedical image analysis challenges*

**Item 8: Submission method**
a) Describe the method used for result submission. Preferably, provide a link to the **submission instructions**.
*Examples:*
- Docker container on the Synapse platform. Link to submission instructions: <URL>
- Algorithm output was sent to organizers via e-mail. Submission instructions were sent by e-mail.

b) Provide information on the possibility for participating teams to **evaluate** their **algorithms before submitting** final results. For example, many challenges allow submission of multiple results, and only the last run is officially counted to compute challenge results.

**Item 9: Challenge schedule**
Provide a **timetable** for the challenge. Preferably, this should include
- the release date(s) of the training cases (if any)
- the registration date/period
- the release date(s) of the test cases and validation cases (if any)
- the submission date(s)
- associated workshop days (if any)
- the release date(s) of the results

**Item 10: Ethics approval**
Indicate whether **ethics approval** is necessary for the data. If yes, provide details on the ethics approval, preferably institutional review board, location, date and number of the ethics approval (if applicable). Add the URL or a **reference to the document** of the ethics approval (if available).

**Item 11: Data usage agreement**
Clarify how the data can be used and distributed by the teams that participate in the challenge and by others during and after the challenge. This should include the explicit **listing of the license** applied.
*Examples:*
- CC BY (Attribution)
- CC BY-SA (Attribution-ShareAlike)
- CC BY-ND (Attribution-NoDerivs)
- CC BY-NC (Attribution-NonCommercial)
- CC BY-NC-SA (Attribution-NonCommercial-ShareAlike)
- CC BY-NC-ND (Attribution-NonCommercial-NoDerivs)

**Item 12: Code availability**
a) Provide information on the **accessibility of the organizers' evaluation software** (e.g. code to produce rankings). Preferably, provide a link to the code and add information on the supported platforms.

b) In an analogous manner, provide information on the **accessibility of the participating teams' code.**

**Item 13: Conflicts of interest**
Provide information related to conflicts of interest. In particular provide information related to **sponsoring/ funding** of the challenge. Also, state explicitly who had/will have **access to the test case labels** and when.

## MISSION OF THE CHALLENGE

**Item 14: Field(s) of application**
State the **main field(s) of application** that the participating algorithms target.
*Examples:*
- Diagnosis
- Education
- Intervention assistance
- Intervention follow-up
- Intervention planning
- Prognosis
- Research
- Screening
- Training
- Cross-phase

**Item 15: Task category(ies)**
State the **task category(ies)**.
*Examples:*
- Classification
- Detection
- Localization
- Modeling
- Prediction
- Reconstruction
- Registration
- Retrieval
- Segmentation
- Tracking

**Suppl 2:** Structured description of a challenge design

*BIAS: Transparent reporting of biomedical image analysis challenges*

**Item 16: Cohorts**
We distinguish between the *target cohort* and the *challenge cohort*. For example, a challenge could be designed around the task of medical instrument tracking in robotic kidney surgery. While the challenge could be based on *ex vivo* data obtained from a laparoscopic training environment with porcine organs (challenge cohort), the final biomedical application (i.e. robotic kidney surgery) would be targeted on real patients with certain characteristics defined by inclusion criteria such as restrictions regarding gender or age (target cohort).

a) Describe the **target cohort**, i.e. the subjects/objects from whom/which the data would be acquired in the final biomedical application.

b) Describe the **challenge cohort**, i.e. the subject(s)/object(s) from whom/which the challenge data was acquired.

**Item 17: Imaging modality(ies)**
Specify the **imaging technique(s)** applied in the challenge.

**Item 18: Context information**
Provide additional **information given along with the images**. The information may correspond ...
a) … directly to the **image data** (e.g. tumor volume).

b) … to the **patient** in general (e.g. gender, medical history).

**Item 19: Target entity(ies)**
a) Describe the **data origin**, i.e. the region(s)/part(s) of subject(s)/object(s) from whom/which the image data would be acquired in the final biomedical application (e.g. brain shown in computed tomography (CT) data, abdomen shown in laparoscopic video data, operating room shown in video data, thorax shown in fluoroscopy video). If necessary, differentiate between target and challenge cohort.

b) Describe the **algorithm target**, i.e. the structure(s)/subject(s)/object(s)/component(s) that the participating algorithms have been designed to focus on (e.g. tumor in the brain, tip of a medical instrument, nurse in an operating theater, catheter in a fluoroscopy scan). If necessary, differentiate between target and challenge cohort.

**Suppl 2:** Structured description of a challenge design

*BIAS: Transparent reporting of biomedical image analysis challenges*

**Item 20: Assessment aim(s)**
Identify the **property(ies) of the algorithms to be optimized** to perform well in the challenge. If multiple properties are assessed, prioritize them (if appropriate). The properties should then be reflected in the metrics applied (parameter 26), and the priorities should be reflected in the ranking when combining multiple metrics that assess different properties.
- *Example 1:* Find liver segmentation algorithm for CT images that processes CT images of a certain size in less than a minute on a certain hardware with an error that reflects inter-rater variability of experts.
- *Example 2:* Find lung tumor detection algorithm with high sensitivity and specificity for mammography images.

Corresponding metrics are listed below (parameter 26).

## CHALLENGE DATA SETS

**Item 21: Data source(s)**
a) Specify the **device(s)** used to acquire the challenge data. This includes details on the device(s) used to acquire the imaging data (e.g. manufacturer) as well as information on additional devices used for performance assessment (e.g. tracking system used in a surgical setting).

b) Describe relevant details on the imaging process/**data acquisition** for each acquisition device (e.g. image acquisition protocol(s)).

c) Specify the **center(s)/institute(s)** in which the data was acquired and/or the **data providing platform/source** (e.g. previous challenge). If this information is not provided (e.g. for anonymization reasons), specify why.

d) Describe relevant **characteristics** (e.g. level of expertise) **of the subjects** (e.g. surgeon)/objects (e.g. robot) involved in the data acquisition process (if any).

**Item 22: Training and test case characteristics**
a) State what is meant by one **case** in this challenge. A case encompasses all data that is processed to produce one result that is compared to the corresponding reference result (i.e. the desired algorithm output).
   *Examples:*
   - Training and test cases both represent a CT image of a human brain. Training cases have a weak annotation (tumor present or not and tumor volume (if any)) while the test cases are annotated with the tumor contour (if any).
   - A case refers to all information that is available for one particular patient in a specific study. This information always includes the image information as specified in *data source(s)* (parameter 21) and may include context information (parameter 18). Both training and test cases are annotated with survival (binary) 5 years after (first) image was taken.

b) State the **total number** of training, validation and test cases.

c) Explain **why a total number** of cases and **the specific proportion** of training, validation and test cases was chosen.

d) Mention **further important characteristics** of the training, validation and test cases (e.g. class distribution in classification tasks chosen according to real-world distribution vs. equal class distribution) and justify the choice.

**Item 23: Annotation characteristics**
a) Describe the **method for determining the reference annotation**, i.e. the desired algorithm output. Provide the information separately for the training, validation and test cases if necessary. Possible methods include *manual image annotation*, *in silico ground truth generation* and *annotation by automatic methods*.

If human annotation was involved, state the **number of annotators**.

b) Provide the **instructions given to the annotators** (if any) prior to the annotation. This may include description of a training phase with the software. Provide the information separately for the training, validation and test cases if necessary. Preferably, provide a link to the **annotation protocol**.

c) Provide **details on the subject(s)/algorithm(s) that annotated** the cases (e.g. information on **level of expertise** such as number of years of professional experience, medically-trained or not). Provide the information separately for the training, validation and test cases if necessary.

d) Describe the **method(s) used to merge multiple annotations** for one case (if any). Provide the information separately for the training, validation and test cases if necessary.

**Item 24: Data pre-processing method(s)**
Describe the **method(s) used for pre-processing** the raw training data before it is provided to the participating teams. Provide the information separately for the training, validation and test cases if necessary.

**Item 25: Sources of error**
a) Describe the most relevant **possible error sources related to the image annotation**. If possible, **estimate the magnitude** (range) of these errors, using inter-and intra-annotator variability, for example. Provide the information separately for the training, validation and test cases, if necessary.

b) In an analogous manner, describe and quantify **other relevant sources of error**.

## ASSESSMENT METHODS

**Item 26: Metric(s)**
a) Define the **metric(s) to assess a property of an algorithm**. These metrics should reflect the desired algorithm properties described in *assessment aim(s)* (parameter 20). State which metric(s) were used to compute the ranking(s) (if any).
- *Example 1:* Dice Similarity Coefficient (DSC) and run-time
- *Example 2:* Area under curve (AUC)

b) **Justify why** the metric(s) was/were chosen, preferably with reference to the biomedical application.

**Item 27: Ranking method(s)**
a) Describe the **method used to compute a performance rank** for all submitted algorithms based on the generated metric results on the test cases. Typically the text will describe how results obtained per case and metric are aggregated to arrive at a final score/ranking.

b) Describe the method(s) used to manage **submissions with missing results** on test cases.

c) **Justify why** the described ranking scheme(s) was/were used.

**Item 28: Statistical analyses**
a) Provide **details for the statistical methods** used in the scope of the challenge analysis. This may include
- description of the **missing data handling**,
- details about the assessment of **variability of rankings**,
- description of any method used to assess **whether the data met the assumptions**, required for the particular statistical approach, or
- indication of any **software product** that was used for all data analysis methods.

b) **Justify why** the described statistical method(s) was/were used.

**Item 29: Further analyses**
Present further analyses to be performed (if applicable), e.g. related to
- **combining algorithms** via ensembling,
- **inter-algorithm variability**,
- **common problems/biases** of the submitted methods, or
- **ranking variability**.